# MODELING URBANIZATION PATTERNS WITH GENERATIVE ADVERSARIAL NETWORKS

*Adrian Albert*[1,2,*], *Emanuele Strano*[1,3], *Jasleen Kaur*[4], *Marta González*[1,2,5]

[1]Massachusetts Institute of Technology, Cambridge, MA 02139
[2]Lawrence Berkeley National Laboratory (LBNL), Berkeley, CA 94720
[3]German Aerospace Center (DLR), Oberpfaffenhofen, Germany
[4]Philips Research U.S.A., Cambridge, MA 02141
[5]University of California, Berkeley CA 94720

## ABSTRACT

In this study we propose a new method to simulate hyper-realistic urban patterns using Generative Adversarial Networks trained with a global urban land-use inventory. We generated a synthetic urban "universe" that qualitatively reproduces the complex spatial organization observed in global urban patterns, while being able to quantitatively recover certain key high-level urban spatial metrics.

## 1. INTRODUCTION

A long-standing question for urban and regional planners pertains to the ability to realistically simulate, by means of explicit spatial modeling, the displacement of urban land-use [1]. Modeling urban patterns has numerous applications, ranging from understanding of urbanization dynamics (which at certain scale of observation follow few physical laws [2]) to inferring future urban expansion to inform policy makers towards a better and more inclusive planning process [3]. Urban models are typically classified in three main categories, land-use/transportation models, cellular automata and agent-based models [4]. These models explicitly locate the urban land-use given the interactions between spatial co-variates like location of services, population density or land price.

However, data on spatial co-variates are difficult and expensive to compile, and are often not available in developing countries where urban growth is more likely to occur.

The recent availability of remote-sensing-based global land-use inventories and the advancements in deep learning methods offer a unique opportunity for pushing the state of the art of spatially-explicit urban models. In this study we propose a spatial explicit model of urban patterns that is based on Generative Adversarial Networks (GANs) [5] trained with very limited spatial information. GANs are a new paradigm of training machine learning models which have shown impressive results on difficult computer vision tasks such as natural images generation [6]. This is a very active area of contemporary machine learning research, whose potential to learn complex spatial distributions has only in the last year started to become better understood in the computational physical sciences literature. For example, recent work has leveraged GANs to generate synthetic satellite images of urban environments [7, 8], de-noise telescope images of galaxies[9], or generate plausible "virtual universes" by learning from simulated data on galaxies [10].

Using a global training samples of $30,000$ cities (urban footprint scenes), we show that a basic, unconstrained GAN model is able to generate realistic urban patterns that capture the great diversity of urban forms across the globe. We see this as a first step towards flexible urban land use simulators for more accurate projections on urbanization in regions where local data is unavailable and difficult to obtain. Next, we outline the basic GAN architecture used (Sec. 2), present experimental results and an empirical validation of the model (Sec. 3). We conclude with key open questions of designing generative models for urban land use analysis (Sec. 4). All code and experiments for this study are available at `https://github.com/adrianalbert/citygan`.

## 2. MATERIALS AND METHODS

### 2.1. Generative adversarial networks (GANs)

Generative adversarial networks (GANs) [5] represent a novel paradigm of training unsupervised machine learning models that learn representations of the input data by training two networks against each other. In the original formulation [5], a *generator* $G$ receives as input a random noise vector $z$, which it transforms in a deterministic way (e.g., by passing it through successive deconvolutional layers if $G$ is a deep CNN) to output a sample $x_{\text{fake}} = G(z)$. The *discriminator* $D$ takes an input $x$ (which can be either real, from an empirical dataset, or synthetically generated by $G$), and outputs the source probability $P(o|x) = D(x)$ that $x$ is either sam-pled from the real distribution ($o =$ real), or produced by $G$ ($o =$ fake). When $G$ is optimal, $x_{\text{fake}}$ is implicitly sampled from the data distribution that $G$ seeks to emulate. This process is summarized schematically in Figure 3*a)*.

*Corresponding author (email: adalbert@mit.edu).

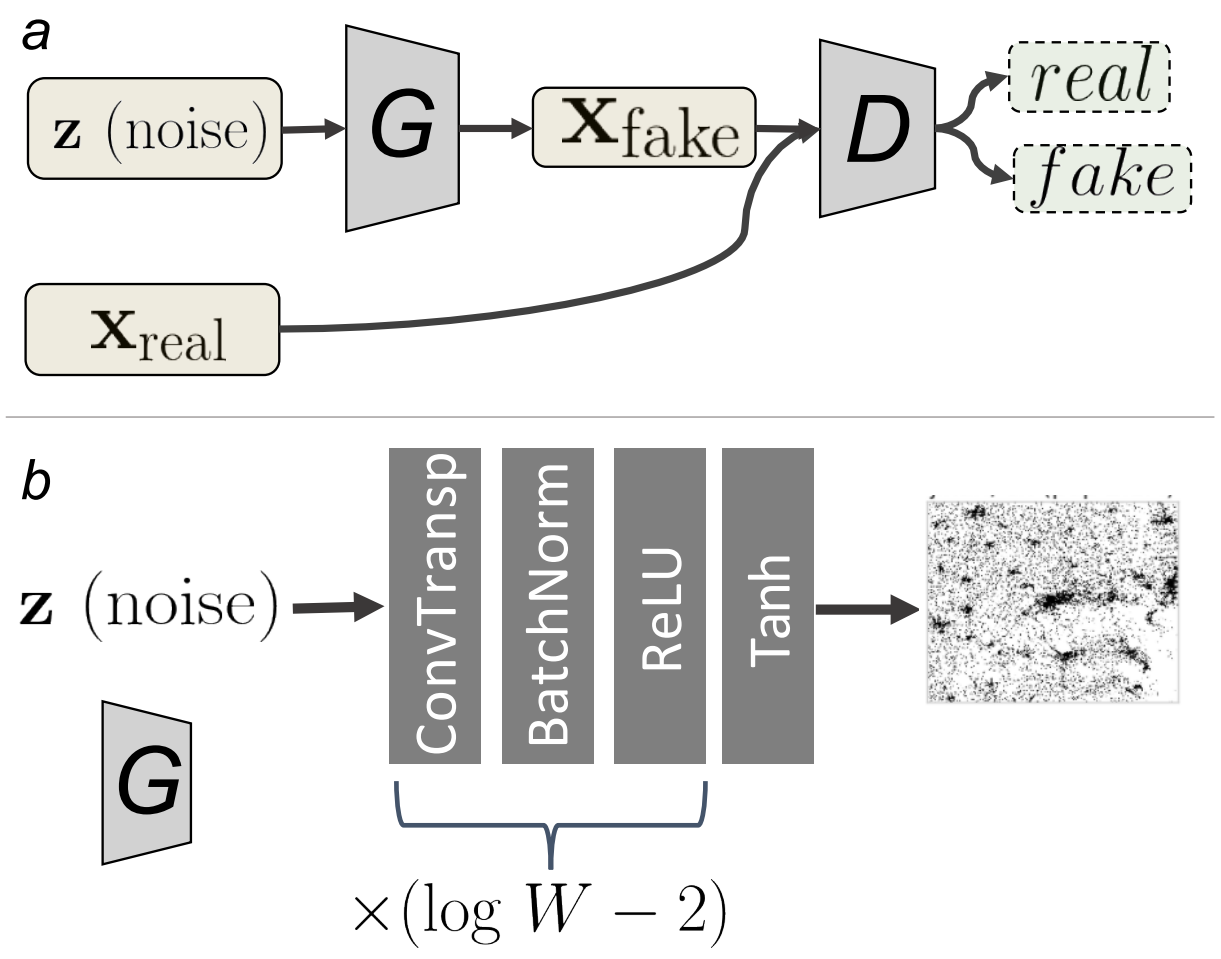

**Fig. 1**. *a)* GAN model architecture following [5]; *b)* the architecture for generator $G$ following [6] is composed of inverse-convolutional, batch normalization, and rectified linear unit (ReLU) layers (the architecture for $D$ is similar).

Both $G$ and $D$ are deep convolutional neural networks parametrized by the weights vectors $\theta_G$ and $\theta_D$. These weights are learned via back-propagation [5] by alternatively minimizing the following loss functions:

$$\theta_D : \mathcal{L}_D = \mathbb{E}_{x\sim p_x}[\log D(x)] + \mathbb{E}_{z\sim p_z}[\log(1 - D(G(z)))], \tag{1}$$

$$\theta_G : \mathcal{L}_G = \mathbb{E}_{z\sim p_z}[\log\,(1 - D(G(z)))] \tag{2}$$

The architectures we used for both $G$ and $D$ follow closely those proposed in [6]. The generator architecture is illustrated in Figure 3*b)*. It is composed of is composed of several convolutional blocks consisting of inverse-convolutional, batch normalization, and rectified linear unit (ReLU) layers, ending in a hyperbolic tangent layer (which applies a $\tanh(\cdot)$ nonlinearity to each element of the generated map). For the discriminator $D$ we used a very similar architecture, with the only differences being leaky ReLU non-linearities instead of the ReLU non-linearities in $G$ and convolutional layers instead of transposed convolutions.

### 2.2. Training sample: built-up areas at global-scale

Here we focused on the simplest, and arguably the most informative spatial feature of cities, which is the presence of built-up areas. To construct a training sample, we used the "Global Urban Footprint" (GUF)[11], an updated global inventory of built-up land at $\sim 12m/px$ resolution. This dataset is published by the German Aerospace Center (DLR) and has been obtained through extensive processing of synthetic aperture radar (SAR) satellite scenes acquired between 2011-2012. We used the built-up footprint of all cities with at least $10,000$ inhabitants (which we estimated by combining population estimates from the LandScan data [12] with city administrative boundaries worldwide from the GADM dataset [13]). For each city, we extracted a sample maps as a square windows of $100 \times 100km$ centered on city center. Fixing a spatial scale of $L = 100km$ results in different image sizes (in pixels) for cities at different latitudes on Earth. We aggregate each extracted map at $750m/px$ resized to $128 \times 128$ pixels. The final training dataset contains $N = 29,980$ binary maps (images) $\mathbf{x}^i$, $i = 1, ..., N$, with $\mathbf{x}^i \in \mathbb{R}^{W\times W}$ and $W = 128$.

## 3. MODEL RESULTS AND VALIDATION

Having trained a generator $G$, we simulated a synthetic "urban universe" of $30,000$ urban maps. Figure 2 illustrates randomly-selected real (left panel) with simulated urban patterns (right panel). At a visual inspection simulations are practically indistinguishable from the real scenes, exhibiting realistic concentrations and spreads of urban masses, including those characteristic of coastal or inland cities. This is in the absence of externally-imposed constraints (e.g., informing the model that water areas cannot be built up). However, aside from qualitative comparisons, it is difficult to quantify the "realism" of a simulated city, since humans have not innate abilities to recognize remote-sensing images of cities (as it is the case in natural images that gave rise to metrics like "Inception score" to quantify GAN performance [14]).

Thus, our validation strategy is to use spatial summary statistics on urban form to compare real against simulate cities. The *average radial profile* $x(d)$ [15] is perhaps one of the most widely-accepted such tools in the urban analysis literature. We compute $x(d)$ by averaging the total amount of built-up area $x$ within rings of width $\Delta d$ at at a distance $d$ from the center (see Figure 3, i.e., values $x(u,v) : (u,v) \in \mathcal{R}(d)$, with $\mathcal{R}(d) \equiv \{(u,v)|(u-u_0)^2 + (v-v_0)^2 > d^2$ and $(u-u_0)^2 + (v-v_0)^2 \leq (d+\Delta d)^2\}$:

$$x(d) \equiv \frac{1}{|\mathcal{R}(d)|} \sum_{(u,v)\in\mathcal{R}(d)} x(u,v) \tag{3}$$

We used the radial profiles in Eq. (3) to determine the polycentric nature of real and simulated scenes via a peak-search algorithm, as illustrated in Fig.3. The peak search algorithm finds points in a univariate profile whose value (peak height) as fraction of maximum is at least $h$, and at a distance from a previously-identified peak of at least $\delta$. We set acceptable values $h = 50\%$ and $\delta = 5\ km$ via experimentation. In Figure 4*a)* we compare the distributions of the number of peaks for all cities on Earth with that for the simulated urban universe. The two distributions show similar form (the $p$-value on a $\chi^2$ test is $\sim 10^{-6}$).

As a further validation, we clustered the radial profiles of real cities and compared to the typical profiles of synthetic ones. To cluster the profiles, we used the `K-Means` algorithm [16]. The results are summarized in Fig. 4. Using a simple fraction of sum-of-squares argument [16], we identify $K = 12$ as the best number of clusters for both real and synthetic scenes. As shown in Fig. 4 *b)* the profile classes are generally very similar as also visible in the panel *c)* where distribution of number of scenes per classes is shown. The distributions are again similar, although for the classes 3 (monocentric cities) and 8 (sprawled patterns) we observe larger differences. We argue that such differences can be due of the sampling strategy which would have favor the abundance of mono-centric urban patterns, while the simulation have been generated regardless the position of the urban core. Note that, by computing average centroids for each of the profile classes, narrower peaks get averaged out. This is an artifact of "measuring" spatial built land use maps in this simple way; indeed, the peak (layer) count and average profile class offer two complementary views on which to compare spatial distributions.

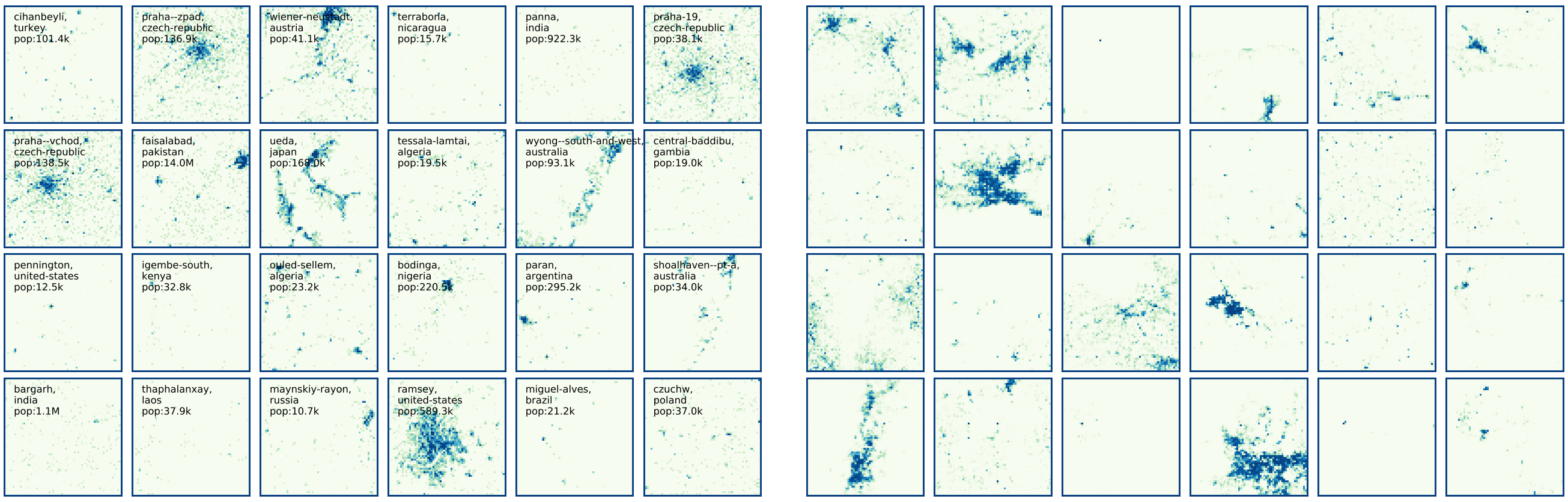

**Fig. 2**. Comparing real urban built land use maps (left) with synthetic maps (right) simulated with a Generative Adversarial Network (GAN). In each case the pixel values are in $[0, 1]$ and represent the fraction of land occupied by buildings.

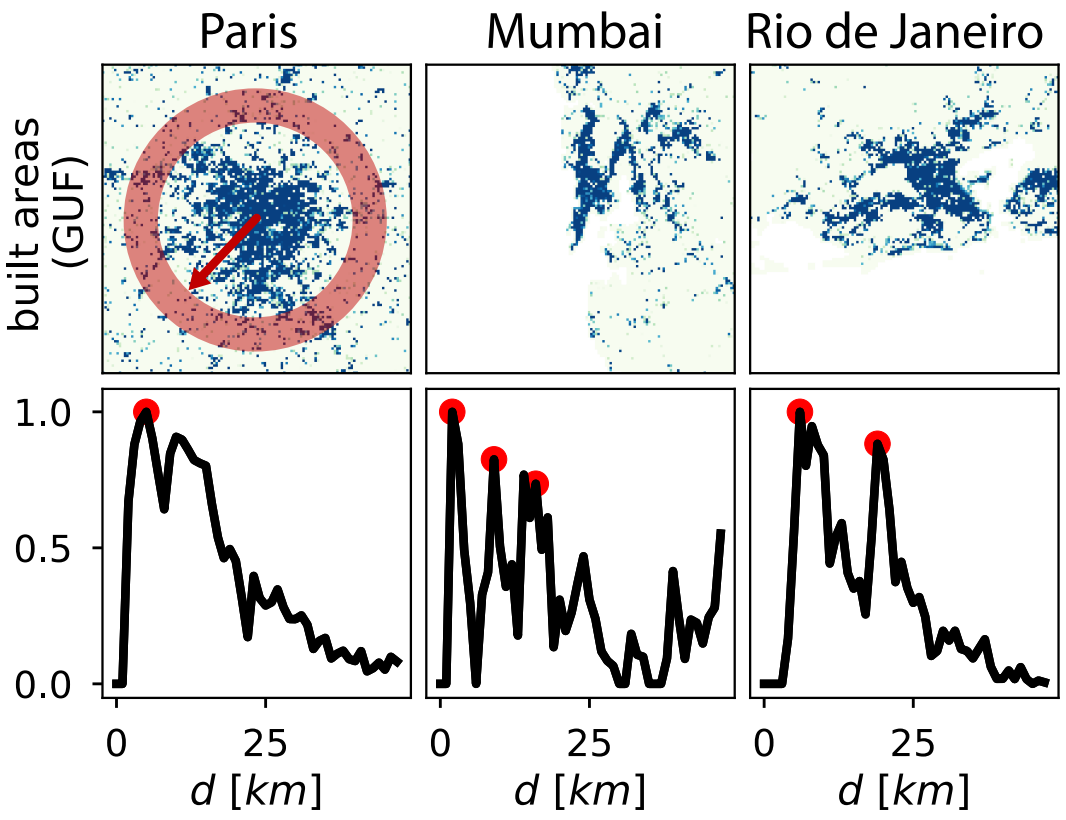

**Fig. 3**. Built land use maps of three example cities (upper row) with their average radial profiles (bottom row). The red dots indicate the peaks detected by the peak-search algorithm.

## 4. DISCUSSION AND CONCLUSIONS

In this study we shown, for the first time, that modern generative machine learning models such as GANs can successfully be used to simulate realistic urban patterns. This is but a start, and despite the impressive results important several open questions still remain. Most of them, as typically for deep-learning (DL) models, pertain to the black-box nature of deep neural networks, which currently lack comprehensive human interpretability and ability for fine-tuned control. We believe, however, that this limitation, which certainly deserves (and gets) attention in the DL literature (e.g., [17]) should not preclude research into their promise to augment existing models using globally-available remote-sensing data.

Important open questions remain: How to evaluate the quality of model output in a way that is both quantitative, and interpretable and intuitive for urban planning analysis? How to best disentangle, explore, and control latent space representations of important characteristics of urban spatial maps? How to learn from both observational and simulated data on cities? In addition, this initial work has only focused on a static snapshot; another area of research is to model city evolution over time using available remote-sensing data (e.g., via the GUF dataset on built land we used here). We plan to address these open questions in on-going work.

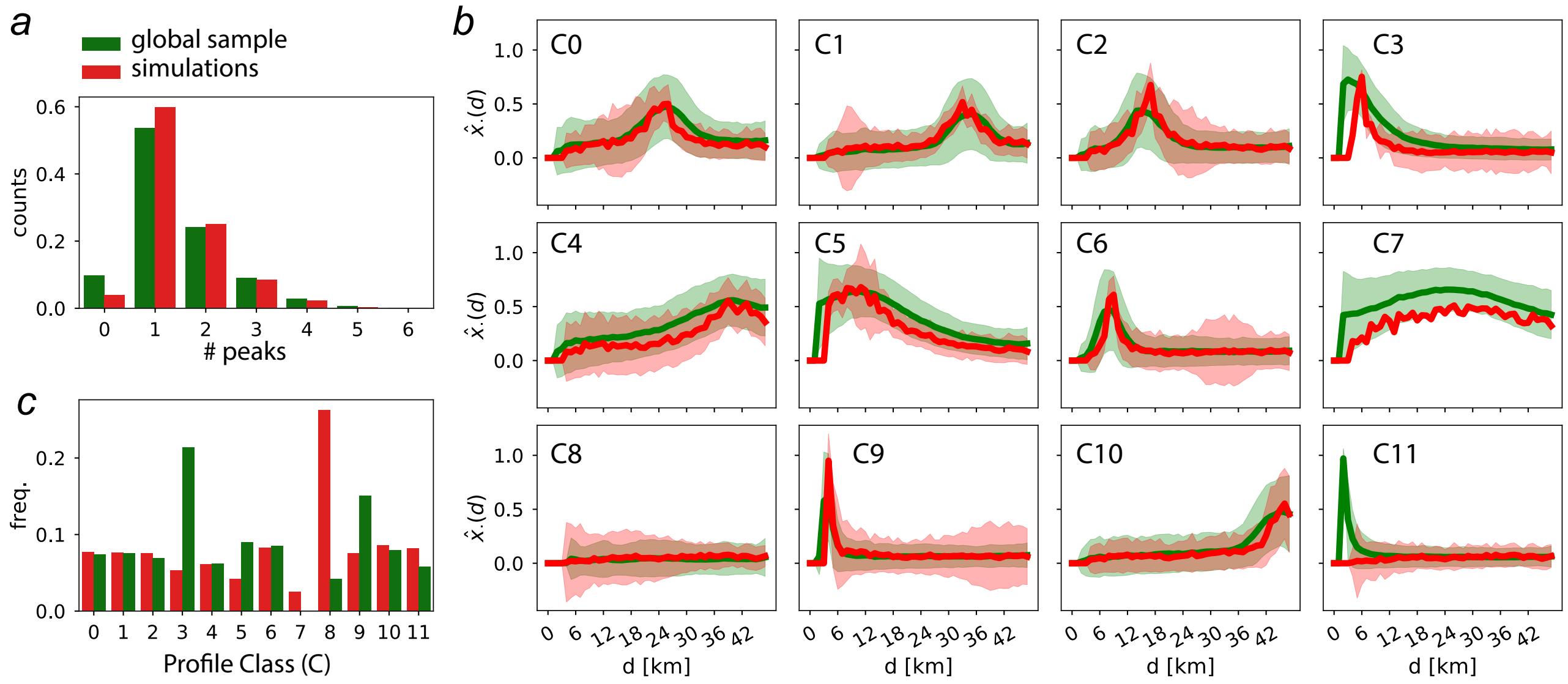

**Fig. 4**. *a)*: the distribution of number of satellite urban centers for real and synthetic cities; c): the distribution of radial profile classes for real and synthetic cities; *b)*: typical radial profiles for real and synthetic cities.